\documentclass{article}
\usepackage[utf8]{inputenc}
\DeclareUnicodeCharacter{0300}{\`{}} % Example for combining grave accent

\usepackage[T1]{fontenc}
\usepackage{authblk}
\usepackage{setspace}
\usepackage[margin=1.25in]{geometry}
\usepackage{float}
\usepackage{graphicx}
\usepackage{subcaption}
\usepackage{amsmath}
\usepackage{lineno}

\usepackage[style=nejm, 
citestyle=numeric-comp,
sorting=none]{biblatex}
\title{Deep Fusion Model for Brain Tumor Classification Using Fine-Grained Gradient Preservation}

\author[1]{Niful Islam}
\author[2$\dag$]{Mohaiminul Islam Bhuiyan}
\author[3$\dag$]{Jarin Tasnim Raya}
\author[2]{Nur Shazwani Kamarudin}
\author[4,5]{Khan Md Hasib}
\author[6*]{M. F. Mridha}
\author[1]{Dewan Md. Farid}

\affil[1]{Department of Computer Science and Engineering, United International University, United City, Madani Avenue, Badda, Dhaka 1212, Bangladesh.}
\affil[2]{Universiti Malaysia Pahang Al-Sultan Abdullah, 26600 Pekan, Pahang, Malaysia.}
\affil[3]{Department of Computer Science \& Engineering, University of Asia Pacific, 74/A Green Rd, Dhaka 1205, Bangladesh.}
\affil[4]{Department of Computer Science and Software Engineering, The University of Western Australia, Perth, WA 6009, Australia.}
\affil[5]{Department of Computer Science and Engineering, Bangladesh University of Business and Technology, Dhaka 1216, Bangladesh.}
\affil[6]{Department of Computer Science, American International University – Bangladesh, Dhaka 1229, Bangladesh.}

\affil[*]{Address correspondence to: firoz.mridha@aiub.edu}
\affil[$\dag$]{These authors contributed equally to this work.}

\date{}

\DeclareUnicodeCharacter{0301}{\'{e}}
\begin{document}

\maketitle

%%%%%% Abstract %%%%%%
\begin{abstract}
Brain tumors are one of the most common diseases that lead to early death if not diagnosed at an early stage. Traditional diagnostic approaches are extremely time-consuming and prone to errors. In this context, computer vision-based approaches have emerged as an effective tool for accurate brain tumor classification. While some of the existing solutions demonstrate noteworthy accuracy, the models become infeasible to deploy in areas where computational resources are limited. This research addresses the need for accurate and fast classification of brain tumors with a priority of deploying the model in technologically underdeveloped regions. The research presents a novel architecture for precise brain tumor classification fusing pretrained ResNet152V2 and modified VGG16 models. The proposed architecture undergoes a diligent fine-tuning process that ensures fine gradients are preserved in deep neural networks, which are essential for effective brain tumor classification. The proposed solution incorporates various image processing techniques to improve image quality and achieves an astounding accuracy of 98.36\% and 98.04\% in Figshare and Kaggle datasets respectively. This architecture stands out for having a streamlined profile, with only 2.8 million trainable parameters. We have leveraged 8-bit quantization to produce a model of size 73.881 MB, significantly reducing it from the previous size of 289.45 MB, ensuring smooth deployment in edge devices even in resource-constrained areas. Additionally, the use of Grad-CAM improves the interpretability of the model, offering insightful information regarding its decision-making process. Owing to its high discriminative ability, this model can be a reliable option for accurate brain tumor classification.
\end{abstract}

%%%%%% Main Text %%%%%%

\section{Introduction}
Brain tumors are abnormal formations of cells that may occur in the brain or in the tissues surrounding the central nervous system (CNS). They may be classified as either cancerous (malignant) or noncancerous (benign). Malignant tumors exhibit fast growth and have the ability to spread to surrounding tissues, whereas benign tumors develop at a slower pace. According to statistical data, almost 70\% of brain tumors are classified as benign, while the remaining 30\% are categorized as malignant \cite{sultan2019multi}. Researchers have identified over 120 distinct forms of brain tumors, including prominent examples such as Gliomas, Meningiomas, and Pituitary Tumors. Gliomas, the predominant kind of brain tumor, arise from glial cells that support and nourish neurons. They constitute 30\% of CNS and 80\% of Glioma tumors are malignant. Meningiomas are benign tumors that grow slowly and originate in the meninges, which are the membranes that surround the brain and spinal cord inside the skull \cite{vankdothu2022brain}. Pituitary tumors arise inside the pituitary glands, which are responsible for hormone regulation and bodily functions and can range from benign to potentially invasive or cancerous. Complications from pituitary tumors can lead to long-term hormonal imbalances and vision problems \cite{kokkalla2021three}. Brain tumors can cause various neurological symptoms, including migraines, epileptic seizures, and cognitive decline. In addition to affecting physical health, they can also affect emotional and psychological well-being. Children with brain tumors may experience delays in social and academic development \cite{irmak2021multi}. Tumors near the sensory processing areas can lead to vision and hearing abnormalities. Medical professionals typically use medical history, physical examinations, and imaging tests such as computed tomography (CT) or MRI scans to diagnose brain tumors \cite{kumar2023human}. Biopsies, which include extracting tiny tissue samples for microscopic analysis, are often required to ascertain the characteristics and severity of the tumor. MRI and CT scans help categorize brain tumors based on their appearance in imaging studies \cite{khan2022accurate}. Precise categorization of brain tumors is essential for developing personalized treatment plans, assessing prognosis, and conducting research to improve treatment outcomes. Neurosurgeons, neurologists, and oncologists combine these classification techniques to determine the most appropriate treatment approach \cite{athisayamani2023feature}. 

Convolutional Neural Networks (CNNs) have become a potent tool for classifying brain tumors in the area of medical image analysis, particularly when working with images from MRI or CT scans. CNNs are a specific kind of deep learning algorithms that are highly regarded for their ability to effectively identify and categorize patterns and classes present in images. These networks operate through multiple layers, including specialized convolutional layers that excel at extracting critical features such as edges and textures from images. Subsequent pooling operations refine these features and preserve essential details while reducing spatial dimensions. CNNs' prowess in image processing is attributable to their utilization of convolution, a technique that filters images to extract pertinent features vital for specific tasks \cite{kurdi2023brain}. One limitation of CNNs is the requirement of large amounts of training data to produce satisfactory performance. Additionally, the training process is often time and resource-intensive. Transfer learning presents a solution to this problem by applying pre-existing machine learning knowledge to similar tasks \cite{asif2023enhanced}. Computer vision models are often trained on large datasets, such as ImageNet. After the model has been trained, it may be used to categorize different sorts of images with little fine-tunings. Pre-trained models achieve very high classification performance by consuming only a few computations. 

This study presents a new fusion architecture to effectively overcome the issue of precise categorization of brain tumors. The proposed architecture extracts features from pre-trained ResNet152V2 and modified VGG16, which are fused and passed through a dual attention module. After fine-tuning, the features were forwarded to XGBoost for the final classification. Notably, we made significant enhancements to the VGG16 architecture to preserve the fine gradients that are essential for the categorization of brain tumors. In the revised design, the features derived from the third, fourth, and fifth blocks were processed and used directly for categorization. Furthermore, we meticulously adjusted the model to ensure that it consumes minimal resources throughout the training process. Additional compression using 8-bit quantization ensured the model was deployable in remote devices. We conducted an assessment of the suggested model using the Fighsare and Kaggle datasets, using multiple evaluation measures. The results demonstrated a higher performance compared to existing solutions. To summarize, this work provides the following significant contributions.  
\begin{itemize}
    \item We have presented a novel architecture by fusing ResNet152V2 and VGG16 for effective brain tumor classification. The VGG16 network employed in the fusion process is altered to preserve the fine gradients that are essential for brain tumor classification. The model is further fine-tuned by replacing the MLP classification head with XGBoost. The improvement has led to an increase in classification accuracy by 6.2\% and a reduction in resource usage, with a size decrease of 37.86 MB.
    \item  The architecture has only 2.8 million trainable parameters, making the training process extremely resource-efficient. The model is further compressed with 8-bit quantization that resulted in a model of size 73.881 MB in size.
    \item The proposed solution incorporates various image processing techniques to enhance the image quality. The resulting images demonstrate enhanced clarity in delineating the tumor regions.
    \item The model has achieved an outstanding accuracy of 98.36\% in classifying three types of brain tumors on the Figshare dataset and 98.04\% accuracy in differentiating tumorous and non-tumorous images on the Kaggle dataset. A comparative analysis with the existing solutions illustrates the superiority of the proposed solution.
    \item To understand the model's decision-making process, we have incorporated explainable AI technologies. According to the analysis, the model effectively identifies the tumor regions, providing insights into its classification mechanisms and enhancing interpretability.
\end{itemize}

The subsequent sections of the paper are structured as follows: Section \ref{backgroundStudy} offers a thorough analysis of prior research. Section \ref{methodology} presents the methodology used in the study. Experimental findings are presented in section \ref{experiment}. Section \ref{conclusion} contains the conclusion and future work.

\section{Related Works}
\label{backgroundStudy}
In recent times, researchers have suggested various techniques for categorizing images of brain tumors. Among them, CNN-based approaches are the most commonly used strategies due to their effectiveness. Noreen et al. \cite{noreen2020deep} conducted an experiment where they fine-tuned the pre-trained Inception-v3 and DenseNet201 models to assess their performance in classifying brain tumors. Their objective was to discover which model outperformed the other. During the fine-tuning phase, the authors eliminated some blocks (i.e. the inception block and dense block) from the lower levels of both models. The features generated from the intermediate inception and dense blocks were then sent through a succession of average pooling, dropout, and fully connected layers before being concatenated and categorized using softmax. According to the experiment, the fine-tuned DenseNet outperformed Inception-v3 by a small margin. High-resolution medical images can play a crucial role in the classification of various diseases. Nevertheless, due to the high expense of acquisition and storage, their widespread utilization may be limited. Mohsen et al. \cite{mohsen2023brain} addressed this issue by employing a Generative Adversarial Network (GAN). The generator produced a 256$\times$256 image using a 64$\times$64 size image as input. The high-resolution image was then classified with ResNext and VGG networks. Asif et al. \cite{asif2022improving} experimented with four pre-trained deep neural networks: Xception, NasNet, DenseNet, and InceptionResNet. After undergoing various data pre-processing and augmentation techniques, Xception outperformed the other three models in the comparison. Shah et al. \cite{shah2022robust} performed fine-tuning on EffecientNetB0 to accurately identify three distinct categories of brain cancers. EfficientNetB0 is a compact image classifier that provides cutting-edge performance by using multi-branch convolution. The authors conducted a comparison between EffecientNetB0 and five state-of-the-art CNN architectures. The experiment suggested the superiority of the proposed solution. Kang et al. \cite{kang2021mri} conducted experiments using several pre-trained deep neural networks to identify the most effective feature extractors. In order to identify the optimal feature extractor, the authors subjected the features to nine machine learning algorithms and evaluated their average performance. Subsequently, the top three best feature extractors (DenseNet, Inception, and ResNeXt) were considered to construct the final ensemble classifier. Khan et al. \cite{khan2022accurate} proposed two CNN architectures for classifying brain tumor images from two different datasets. On the Havard Medical Dataset, the model employed a pre-trained VGG16 and for the Fighshare dataset, the authors proposed a 23-layer CNN architecture. Owing to the high resolution of the input image, the model consumes more than twice the computation time compared to its similar research works.

Researchers have proposed many shallow CNN architectures for brain tumor classification because of the significant computational cost of CNN models. Classifiers with four blocks of CNN provided the highest level of efficiency. Montaha et al. \cite{montaha2022timedistributed} introduced a specialized CNN architecture consisting of four blocks, in which the convolutional and pooling layers were combined with Long Short-Term Memory (LSTM). Through a comprehensive ablation analysis, the suggested model attained an accuracy of 98.90\% on a merged dataset, resembling the prior approach. Additionally, Musallam et al. \cite{musallam2022new} introduced a customized CNN model with a feature extractor consisting of four blocks. The first two blocks consisted of two convolutional layers, whereas the latter two blocks were composed of three convolutional layers apiece. Batch normalization was used after each convolutional layer to accelerate the convergence. Additionally, a 2$\times$2 max-pooling layer was included at the end of the block to decrease the spatial dimension. Amin et al. \cite{amin2020brain}, introduced a novel CNN model consisting of four layers. The model incorporated a Discrete Wavelet Transform (DWT) to merge four different kinds of MRI data into a single image. After the fusion procedure, the authors used a global thresholding strategy to partition the tumor area. The segmented area was then designated as the area of Interest (ROI), which was used to precisely identify the tumor location in the original image. Subsequently, the annotated images were passed through the CNN model for classification. The authors assessed the proposed architecture using five publicly accessible brain tumor datasets and determined that the fused images demonstrated superior performance compared to classification based on a single images. Khairandish et al. \cite{khairandish2022hybrid} introduced a convolutional neural network (CNN) consisting of only two convolutional layers. The dataset was first segmented by the system using threshold-based segmentation. The segmented image was then passed through the CNN feature extractor, which consisted of two 5$\times$5 convolutional layers and two 2$\times$2 pooling layers. The features obtained from the Convolutional Neural Network (CNN) were ultimately categorized using a Support Vector Machine (SVM) classifier. The experiment illustrated that the integration of the SVM increased CNN's classification performance by 1\%. Diaz et al. \cite{diaz2021deep} proposed a three-stream CNN architecture that extracted multi-scale features using only two convolutional layers. The first stream extracted large features with 11$\times$11 convolutional blocks, and the second stream incorporated 7$\times$7 convolutional blocks to capture median scale features. Finally, the last stream consisted of 3$\times$3 convolutional layers to detect small features. The features extracted from all streams were concatenated and classified using a Multi-Layer Perceptron (MLP) head. 

Transformers, initially designed for natural language processing, have now been widely used for image recognition tasks. Some well-known transformer architectures have gained attention due to their superiority over traditional CNN architectures in various image classification challenges \cite{liu2021swin, wang2021pyramid}. Due to their high classification performance, researchers have integrated vision transformers for brain tumor classification. Tummala et al. \cite{tummala2022classification} ensembled four vision transformer (ViT) models for effective brain tumor classification. Among the four ViT models, two of them were ViT Base (B/16 and B/32), and two others were ViT Large (L/16 and L/32). Traditionally ViT models have a huge number of parameters, especially ViT Large models, which boast over three times as many parameters as ViT Base. Furthermore, due to the self-attention mechanism, where each patch must attain every other patch, the computation becomes quadratic, making ViT relatively slow. Therefore, the proposed ensemble approach faces challenges where computational power is moderate. Ferdous et al. \cite{ferdous2023lcdeit} presented a solution to the quadratic time consumption of self-attention by replacing the attention mechanism with external attention, which is of linearly complex. The authors trained the modified transformer through knowledge distillation where a CNN model was considered as the teacher model and the proposed transformer as the student model. Since the feature extraction mechanisms of CNN and transformers were different, some researchers have combined CNN and transformers to attain better accuracy. Aloraini et al. \cite{aloraini2023combining} presented a two-stream architecture where the first stream was comprised of a transformer and the second stream was a CNN. For the CNN stream, the authors employed a pre-trained DneseNet201. Due to the local and global feature extraction with the CNN and transformer, respectively, the proposed solution achieved high classification performance.

\begin{table}[!ht]
    \centering
    \caption{Comparative analysis of existing works.}
    \begin{tabular}{ p{2cm} p{2.6cm} p{4cm} p{4cm}}
    \hline
        Paper & Method & Contribution & Limitation \\ \hline
        Noreen et al. \cite{noreen2020deep} & Inception-v3, DenseNet201 & Comperison between two state-of-the-art methods.  & Lack of widespread utilization.  \\ 
        Mohsen et al. \cite{mohsen2023brain} & GAN, RexNext, VGG & Integrated generative networks to produce new images. & Lacks interpretability of the decision-making process. \\ 
        Asif et al. \cite{asif2022improving} & Xception, NasNet, DenseNet, and InceptionResNe & Comparison between four state-of-the-art methods. & Classification accuracy drops when the number of training samples is less.  \\ 
        Shah et al. \cite{shah2022robust} & EffecientNetB0 & Achieves an outstanding classification accuracy. & Only classifies tumors and non-tumors images. \\ 
        Kang et al. \cite{kang2021mri} & DenseNet, Inception, and ResNeX & Present a novel architecture that classifies tumor and normal images. & The model is extremely resource intensive. \\ 
        Khan et al. \cite{khan2022accurate}  & VGG16, Custom 23-layer model & Presented two separate architectures for two different datasets. & High-resolution images result in more computation; and relatively low classification accuracy. \\ 
        Montaha et al. \cite{montaha2022timedistributed} & Custom CNN-LSTM based model & Presents a novel architecture for brain tumor classification. & The integration of LSTM makes the model slow. \\ 
        Amin et al. \cite{amin2020brain} & ViT, DenseNet & Achieves a very high classification accuracy. & Consumes high computational resources.  \\ \hline
    \end{tabular}
    \label{tab:rw-comp}
\end{table}

A comparative analysis of the presented studies with their major contributions and limitations is presented in Table \ref{tab:rw-comp}. According to the analysis, several studies have been conducted to classify brain tumors effectively. Nonetheless, there remains a gap in high-performance brain tumor classifiers that are resource-efficient for training. Moreover, most studies lack the interpretability of the classifiers which leaves the model's decision-making process unknown. Therefore, we have proposed a high-performing image classifier that has only 2.8M trainable parameters and requires only 50 epochs of training. Moreover, we employed Grad-CAM to spot the areas where the model is focused, aiming to enhance the model's reliability in situations involving domain shifting.

\section{Methodology}
\label{methodology}
The proposed solution comprises three major steps: data pre-processing, model construction, and model quantization. Section \ref{sec:data-preprocess}, \ref{sec:model-construct}, and \ref{sec:model-quantization} illustrate these steps in detail.

% \begin{figure}[h]
%     \centering
%     \includegraphics[height=8cm,width=0.8\linewidth]{FIGURES/sample.png}
%     \caption{A sample of the fighshar dataset. The dataset encompasses three types of tumor images from three distinct viewpoints.}
%     \label{fig:sample}
% \end{figure}

\begin{figure*}[h]
\centering
\begin{subfigure}{0.48\textwidth}
    \includegraphics[height=6cm, width=6.8cm]{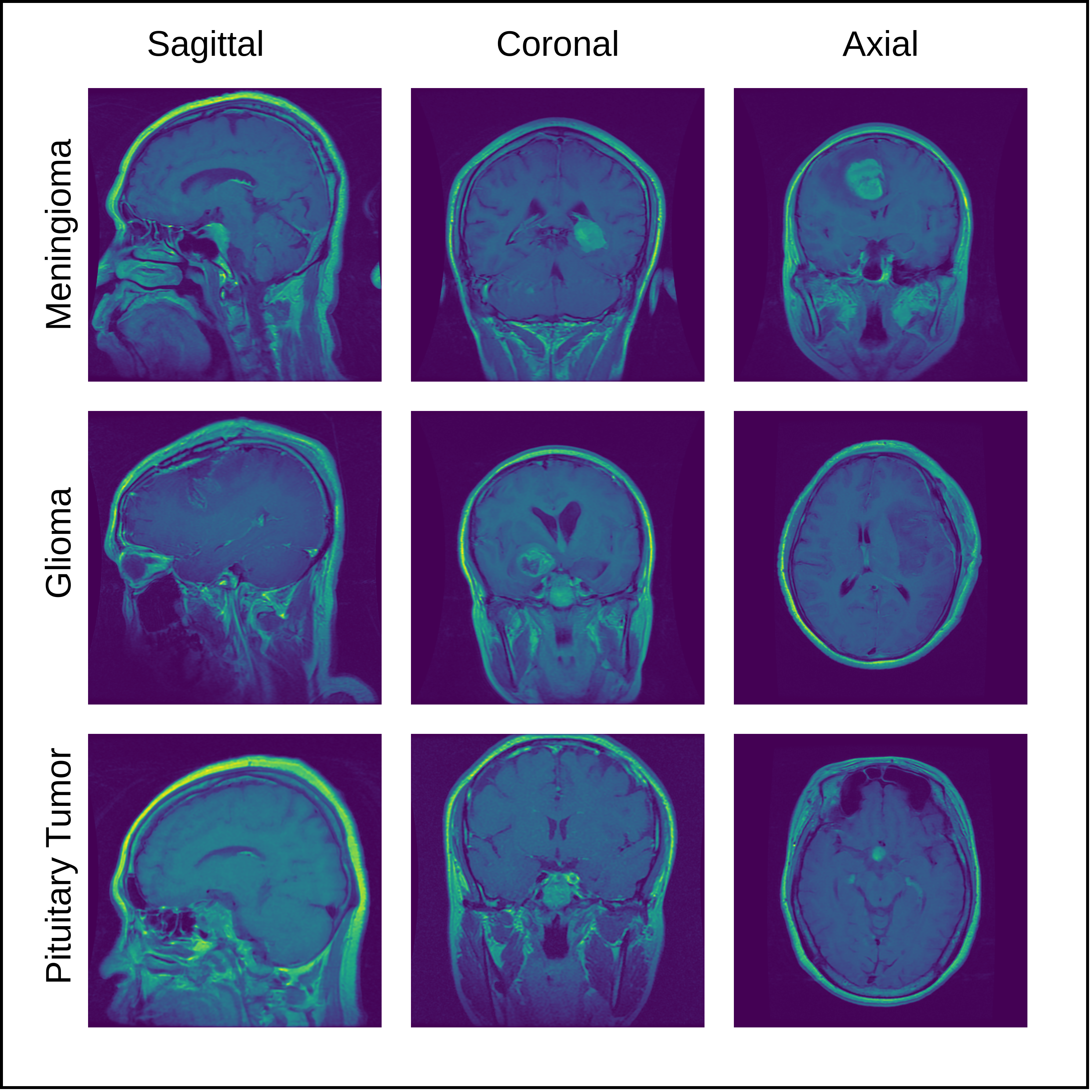}
    \caption{A sample of the Fighshar dataset.}
    \label{fig:sample}
\end{subfigure}
\hfill
\begin{subfigure}{0.48\textwidth}
    \includegraphics[height=6cm, width=6.8cm]{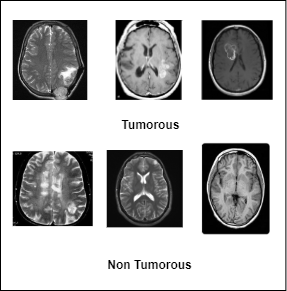}
    \caption{A sample of the Kaggle dataset.}
    \label{fig:tumor}
\end{subfigure}

\caption{A sample of the Figshare and Kaggle dataset.}
\label{fig:data-sample}
\end{figure*}

\subsection{Data Pre-processing}
\label{sec:data-preprocess}
\begin{figure*}[h]
    \centering
    \includegraphics[height=8.7cm,width=1\linewidth]{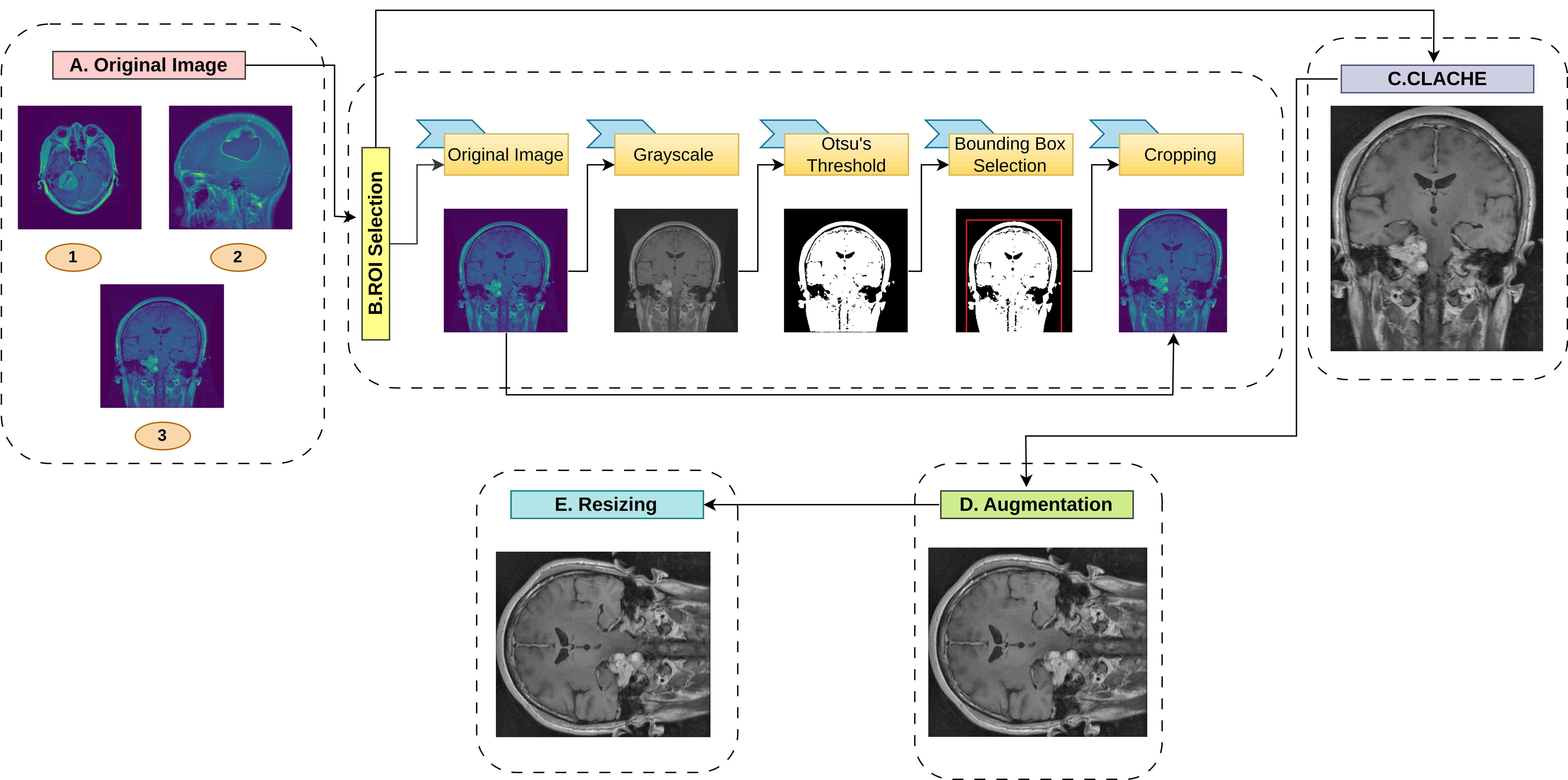}
    \caption{The four-stage data preprocessing technique employed in the system.}
    \label{fig:preprocessing}
\end{figure*}

We have leveraged two publicly available datasets for conducting the study. The first dataset is the Figshare dataset \cite{Cheng2017} that comprises 3064 T1-weighted brain tumor images from three classes. Collected from 233 patients, the slightly imbalanced dataset consisted of 708, 1426, and 930 images of meningiomas, gliomas, and pituitary tumors respectively. The images of that dataset are collected from three distinct viewpoints. The second dataset employed is commonly known as the Kaggle dataset \cite{kaggle_brain_mri} which consists of 253 MRI images. 98 of those images are categorized as non tumorous while the rest of the images are tumorous. Unlike the Figshare dataset, the images of the Kaggle dataset are collected from one viewpoint. Figure \ref{fig:data-sample} shows a sample of the datasets.
% \begin{figure}[h]
%     \centering
%     \includegraphics[height=8cm,width=0.8\linewidth]{FIGURES/tumor.png}
%     \caption{A sample of the Kaggle dataset.}
%     \label{fig:tumor}
% \end{figure}
In the data preprocessing stage, the dataset was first split into three non-overlapping training(70\%), validation(10\%), and test(20\%) sets. The subsequent four steps are the region of interest (ROI) selection, adaptive histogram equalization, data augmentation, and resizing. The details of these processes are presented in Section \ref{sec:roi}, \ref{sec:ahe}, \ref{sec:augment}, and \ref{sec:resizing} respectively. Figure \ref{fig:preprocessing} holds a diagrammatic overview of the preprocessing stage.

\subsubsection{ROI Selection}
\label{sec:roi}
As presented in Figure \ref{fig:sample} and \ref{fig:tumor}, the images have a noticeable dark background that does not contribute to the classification process. Consequently, eliminating these portions enhances the accuracy of classification. In order to choose the region of interest (ROI), the images were first segmented using Otsu's thresholding technique. This image segmentation approach utilizes a histogram to calculate the optimal threshold value by minimizing the intra-class variance. This technique often utilizes grayscale images and produces a binary image \cite{islam2023toward}. The threshold value is determined using Equation \ref{eq:otsu}. Here, $k_{\text{Otsu}}$ represents the predicted threshold value for Otsu's thresholding, whereas $k$ denotes the possible threshold candidates. The value of $k$ lies from 0 to 255. The variables $q_1(k)$ and $q_2(k)$ represent the probabilities of the background and foreground pixels, respectively. Similarly, $\sigma_1^2(k)$ represents the variance of the background pixels, whereas $\sigma_2^2(k)$ represents the variance of the foreground pixels.
\begin{equation}
\label{eq:otsu}
    k_{\text{Otsu}} = \arg\min_{k=0}^{k=255} \left[ q_1(k) \cdot \sigma_1^2(k) + q_2(k) \cdot \sigma_2^2(k) \right]
\end{equation}

Once the threshold is calculated, the intensity values above the threshold are shown in white, while the ones below the threshold are indicated in black. Given its high efficacy, this algorithm has gained extensive use in the categorization of brain tumors \cite{nyo2022otsu}.

To apply Otsu's thresholding, RGB MRI images were first converted into grayscale images. The process of grayscale conversion is presented in Equation \ref{eq:gray}. In this equation, $I_g$ represents the grayscale image while $I_R$, $I_G$, and $I_B$ illustrate the Red, Green, and Blue channels of the image respectively. The grayscale conversion is a linear multiplication of each pixel value with a specific constant.

\begin{equation}
\label{eq:gray}
    I_g = 0.299 \cdot I_R + 0.587 \cdot I_G + 0.114 \cdot I_B
\end{equation}

Subsequently, the grayscale images were segmented using Othu's threshold. A bounding rectangle is drawn on the segmented white region which illustrates the region of interest. Eventually, the rectangular portion of the original RGB image is cropped to select the ROI.

\subsubsection{Adaptive Histogram Equalization}
\label{sec:ahe}
Due to various factors such as variations in tissue thickness, illumination during imaging, or shadows caused by tumors, images in medical imaging, including images of brain tumors, frequently have uneven lighting. Important details in an image may be difficult to visualize because of these variations. Contrast Limited Adaptive Histogram Equalization (CLAHE) can be employed to address this issue. This image processing algorithm enhances the contrast and visibility of details in an image. Standard histogram equalization operates on the entire image which can degrade the quality of a brain tumor image because the tumor portion is a small area of the image. Therefore CLAHE is leveraged to divide the image into small tiles. Histogram equalization was applied within each tile. To prevent over-enhancement and noise amplification, the contrast enhancement of each tile is limited. To prevent noticeable artifacts at the tile boundaries, CLAHE employs a sliding window approach with overlapping tiles. As the tumor lesion contributes a small portion and CLAHE operates with smaller portions of the image, it greatly improves the image quality highlighting the tumor region. In the proposed solution, the tiles are of size 8$\times$8 in size and the contrast limit for localized changes is 2.0.

\subsubsection{Augmentation}
\label{sec:augment}
Since the dataset had a limited number of samples, the training set was augmented to increase the robustness and reduce the overfitting of the model. Six real-time data augmentation techniques were applied as illustrated below. 
\begin{itemize}
    \item Random Rotation: A 40-degree clockwise or counterclockwise rotation is applied at random. 
    \item Height Shift: Random height shift is applied by a maximum of 20\%.
    \item Width Shift: Random width shift is applied by a maximum of 20\%.
    \item Shear Transform: A shear transformation was employed on the input images, utilizing a maximum shear angle of 0.2 degrees.
    \item Zooming: Some random images are zoomed with a zooming range of 20\%.
    \item Random Flipping: Some images are randomly flipped horizontally or vertically. 
\end{itemize}

\subsubsection{Resizing}
\label{sec:resizing}
Resizing is a commonly used data preprocessing technique, particularly in the context of transfer learning. Generally, the state-of-the-art models employ ImageNet \cite{deng2009imagenet} for training the model.  Although the images in ImageNet are of size 469$\times$387, they are mostly resized to 255$\times$255 or 224$\times$244 for architectural compatibility and resource efficiency \cite{magdy2023performance}. In order to fully utilize the advantage of transfer learning, we have also resized the images to 224$\times$244 which makes the model compatible with the pretrainded classifier. We have leveraged the nearest neighbor interpolation algorithm for resizing. The algorithm resizes images by selecting the nearest pixel value from the original image to determine the color of each pixel in the resized image.

\subsection{Model Construction}
\label{sec:model-construct}
\begin{figure*}[h]
    \centering
    \includegraphics[height=7cm,width=0.9\linewidth]{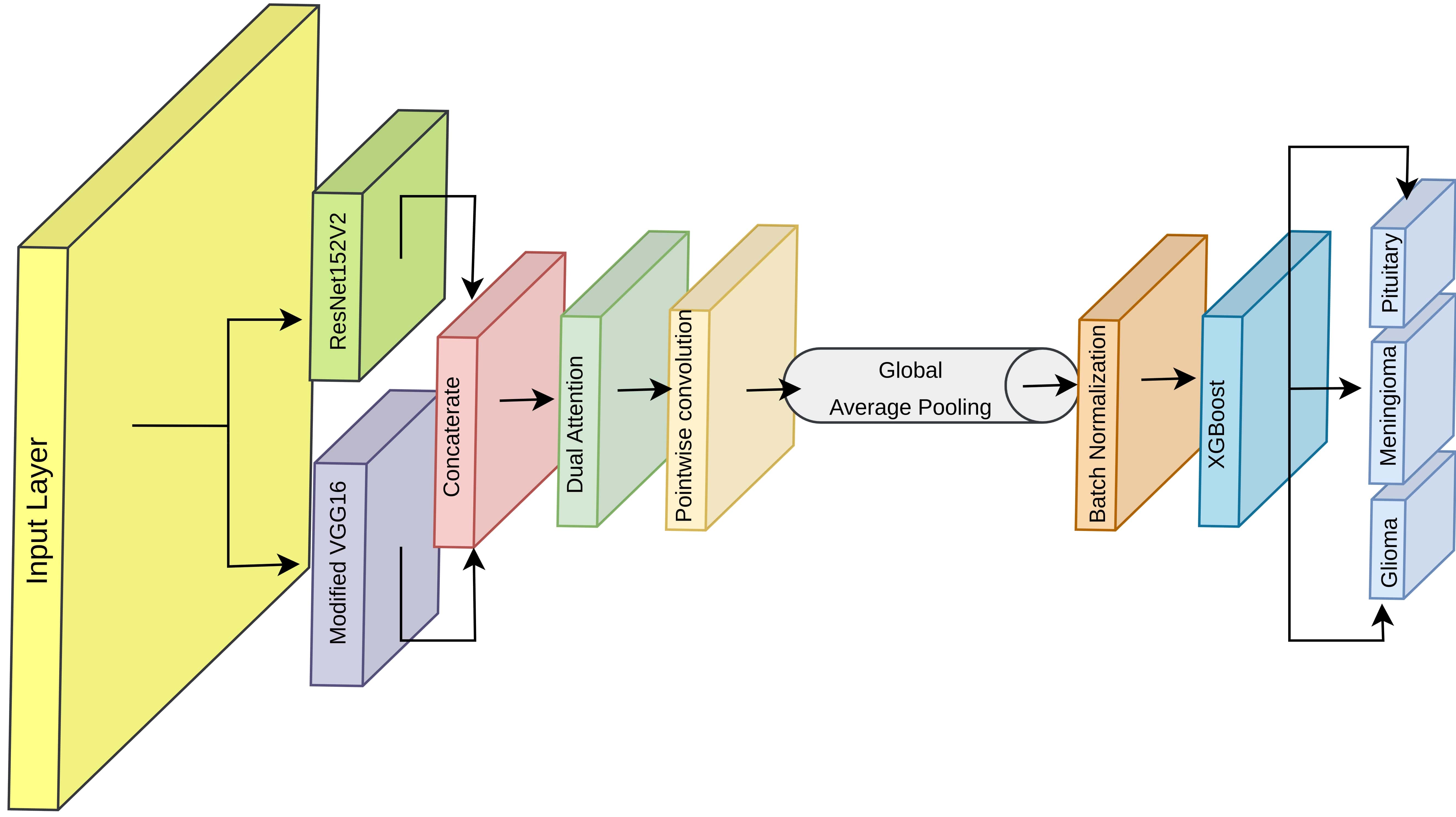}
    \caption{The proposed fusion architecture for brain tumor classification.}
    \label{fig:model}
\end{figure*}

The fusion model, presented in Figure \ref{fig:model}, consists of two advanced image classifiers, namely ResNet152V2 and VGG16. The features obtained from the pre-trained feature extractors are combined and then sent to an attention module to prioritize the important features. Upon completion of the fine-tuning process, the model was quantized and prepared for classification. The following sections provide an elaborate synopsis of the proposed classifier. 

\subsubsection{ResNet152V2}
ResNet152V2 is an improvement over the original ResNet architecture. Traditional sequential CNNs architectures face the vanishing gradient problem, where the gradients of the loss function with respect to the network parameters become extremely small as they are backpropagated through numerous layers. This problem can make it difficult to train deep neural networks, which can cause learning to converge slowly or even stop altogether. ResNets provided a solution to this issue by incorporating skip connections, also known as residual connections \cite{he2016deep}. In this network, the features extracted from the previous block are passed on to the next block. This addresses the vanishing gradient problem while ensuring feature reuse, thereby making the construction of deep neural networks possible. Several versions of ResNet have been developed since its introduction in 2016. ResNet152V2, which is composed of 152 layers, is an improvement over the base ResNet architecture. This model integrates batch normalization which accelerates up the training process \cite{hwooi2022deep}. Moreover, a deeper bottleneck was integrated to extract more complex features while maintaining the complexity manageable. 

\subsubsection{Modified VGG16}
Visual Geometry Group 16, commonly known as VGG16, is a 16-layer convolutional neural network used for various image recognition tasks. Developed in 2014, this deep learning architecture has gained popularity due to its simplistic architecture and high performance. With a stack of convolutional layers followed by max-pooling layers, VGG16 achieved state-of-the-art performance. However, since VGG16 architecture does not integrate any residual connections, some small or fine-grained features tend to be oversmoothed or removed. This characteristic becomes especially important when processing images of brain tumors because the tumor region typically makes up a small portion of the overall image. Accurate tumor detection and characterization depend heavily on recognizing and maintaining minute details. Therefore, we present a novel architecture that addressed this challenge. With minimal fine-tuning, our method passes the features extracted from the third, fourth, and fifth convolutional blocks directly for classification purposes. This method enables us to extract and store essential fine-grained data, thereby enhancing the performance of the model in the accurate localization of tumor regions within brain images. 

\begin{figure*}[h]
    \centering
    \includegraphics[height=7cm,width=0.9\linewidth]{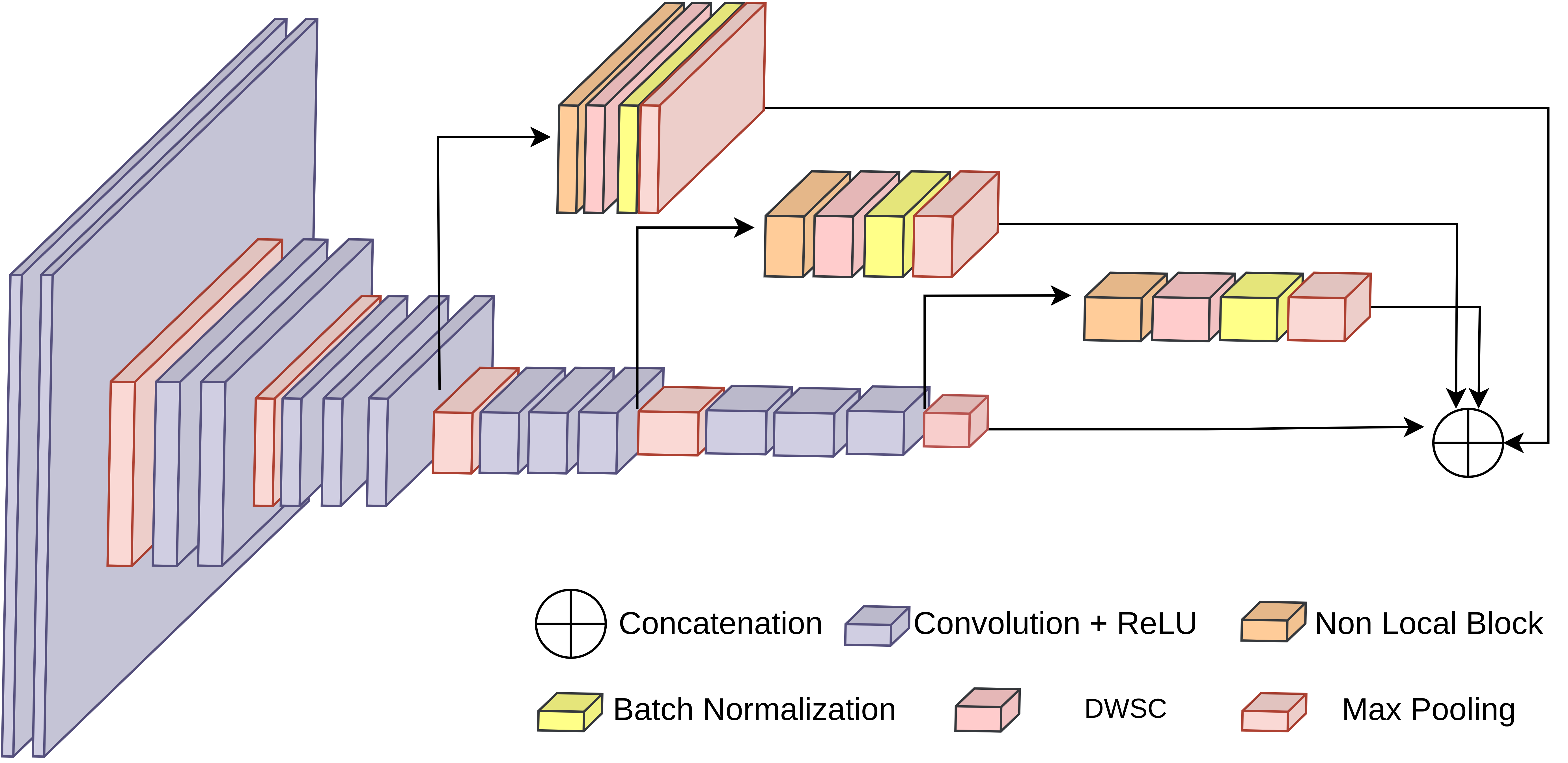}
    \caption{Modified VGG16 network used in the fusion model.}
    \label{fig:modifiedvgg}
\end{figure*}
As presented in Figure \ref{fig:modifiedvgg}, features extracted from the third, fourth, and fifth blocks pass through a series of non-local blocks, depth-wise separable convolution (DWSC), batch normalization, and a max pooling layer. Because the pre-trained VGG16 is trained on ImageNet, this network might extract some irrelevant features that are attended by non-local blocks \cite{wang2018non}. A non-local block attention mechanism is critical in deep learning architectures, particularly convolutional neural networks (CNNs), which are designed to capture long-range dependencies within input feature maps.  This attention mechanism computes the similarity scores between all positions in the feature map. This score is then integrated into the features map allowing the network to capture global dependencies and, boosting the network's generalization ability. Following the non-local block, a DWSC is integrated to effectively reduce the feature map without increasing the computational cost. The DWSC block, composed of depthwise and pointwise convolutions, takes 256, 512, and 512 feature maps from the third, fourth, and fifth blocks respectively, and convert them into feature maps of size 128 each. After DWSC, batch normalization is applied for faster convergence followed by a max pooling layer to match the feature dimensions of the subsequent layers. Although the modification in the fifth (final) block might appear redundant, it helps attain the features while keeping the original feature extraction untouched.

\subsubsection{Fine Tuning}
The fine-tuning process involved fusing the features from ResNet152V2 and modified VGG16. Instead of relying on a single classifier, combining different feature extractors with a heterogeneous architecture produces a significantly better performance. Because the pretrained image classifiers are trained on ImageNet, which represents a different domain,  adopting an attention module after the feature extraction stage can help generalize better in the medical domain, which differs in terms of image characteristics, textures, and structures \cite{kong2022classification}. Therefore, the fusion model integrates a dual attention module after the feature fusion stage. Dual attention is a complex attention module that integrates both channel and spatial attention \cite{fu2019dual}. Let the input features be denoted by $X$ and the height and width of the features be $H$ and $W$ respectively. The channel attention mechanism can be explained using Equation \ref{eq:channel-attention}. To obtain the global description $U$, the global average pooling of each layer was performed. Subsequently, the global description is multiplied by $W_1$, the weight matrix of the first dense layer which is then passed through the ReLU activation function. Similarly, the feature vector is multiplied by $W_2$, the second weight matrix and passed through the sigmoid activation function to obtain the channel attention weight $A_c$. Because the range of the sigmoid funciton is [0, 1], each channel is prioritized based on its importance to the output, with 0 being irrelevant and 1 being fully relevant. Finally, the attention weight was reshaped to match the input feature size and multiplied with it to obtain the channel prioritized output.

\begin{equation}
\label{eq:channel-attention}
\begin{aligned}
U &= \frac{1}{H \cdot W} \sum_{i=1}^{H} \sum_{j=1}^{W} X(i, j, :) \\
A_c &= \sigma\left(\text{ReLU}(UW_1)W_2\right) \\
A_c &= \text{Reshape}(A_c, (1, 1, C)) \\
X_c &= X \cdot A_c
\end{aligned}
\end{equation}

Along with channel attention, spatial attention is calculated simultaneously. In this process, as presented in Equation \ref{eq:spatial-attention}, the spatial attention weights are calculated by applying a 1$\times$1 convolutional layer with sigmoid activation to the input feature maps which are also multiplied by the input features. 

\begin{equation}
\label{eq:spatial-attention}
\begin{aligned}
A_s &= \sigma(W_s \ast X) \\
X_s &= X \cdot A_s
\end{aligned}
\end{equation}

After calculating the of channel and spatial attention, the refined features are summed and returned as output. Equation \ref{eq:sa+ca} illustrates the combination process.

\begin{equation}
\label{eq:sa+ca}
\begin{aligned}
X_{\text{attention}} &= X_c + X_s
\end{aligned}
\end{equation}

On the refined features returned from the dual attention block, a 1$\times$1 convoluton, also known as pointwise convolution, was applied to reduce the feature size. The pre-trained ResNet152V2 returned 2,048 features, whereas the modified VGG16 returned 896 features, which adds up to a total of 2,944. Pointwise convolution converts the large feature size to 128 features only, which results in a significant reduction in the trainable parameters. To convert the 2D feature maps into 1D feature vectors, the architecture integrates Global Average Pooling (GAP) instead of flattening which returns only one value computing the average of each feature map. Although only a single value may be considered insignificant in reflecting the entire feature map, GAP provides an immense improvement over flattening in terms of resource consumption. For comparison, flattening converts each feature map to a feature vector of size 196 (feature map size 14$\times$14), where the GAP returns feature vectors of size 1. The feature vectors then follow a batch normalization layer. 

Initially, after batch normalization, a shallow Multi-Layer Perceptron (MLP) is integrated to train the weights of the trainable parameters including the dual attention block, modified VGG16, and other layers. However, because of the limited number of samples, the MLP failed to produce satisfactory results. Additionally, the fully connected neural network consumed over 37.86 MB of memory which is a concern for deploying the model in a resource-constrained environment. Therefore, the MLP layer was replaced with an XGBoost classifier of size 0.72 KB for final prediction. XGBoost, which is short for Extreme Gradient Boosting, is a widely used machine learning algorithm that is renowned for its outstanding performance in supervised learning. It is an ensemble method that aggregates the predictions of various decision trees to obtain reliable and accurate predictions. By adding new decision trees that rectify mistakes in older treees, XGBoost sequentially improves decision trees. To efficiently handle a variety of datasets, regularization, gradient boosting techniques, and sophisticated features such as weighted quantile drawing are used. Owing to its effectiveness, speed, and capacity to manage missing data, XGBoost has become a considerable option for challenges and practical applications across numerous industries. The model, with 2.8 million trainable parameters, was trained on the categorical cross-entropy loss function and Adamax optimizer for 50 epochs only. 

In summary, the proposed system incorporates various data preprocessing techniques to enhance the image quality. Following data pre-processing, a novel classifier was constructed by fusing ResNet152V2 and VGG16. Due to the oversmoothing problem of VGG16, the model was slightly fine-tuned to keep the essential features intact. Finally, the combined features are classified using XGBoost. The system also employs various techniques to maintain the training resource consumption within an acceptable range without compromising the performance.  

\subsection{Model Quantization}
\label{sec:model-quantization}
In TensorFlow, deep learning models typically have 32-bit precision, which indicates that the weights, biases, and activation values are stored in 32-bit floating-point numbers. While higher bit precision ensures greater precision, lowering the number of bits allocated to each value results in a more compact model, which is a major concern in resource-constrained applications. Model quantization is a technique that systematically converts the weights, biases, and activation values of a neural network from its original floating-point format to a lower-precision format, ensuring minimum or no performance degradation. For compressing the proposed model without compromising the performance, we have leveraged 8-bit quantization that compresses the feature extractor by a factor of 4. The careful compression prevented over-quantization, which resulted in no performance loss. We compressed the feature extractor, which initially had a size of 289.45 MB. After quantization, the feature extractor was reduced to 73.88 MB. Following the feature extractor, the XGBoost classifier remained at a size of 0.72 KB, undergoing no compression. Overall, the full classifier is reduced to a size of less than 73.881 MB through quantization.

\section{Experiment} 
\label{experiment}
This section presents the outcomes derived from the proposed model, along with a thorough comparison with the existing research. 
\subsection{Experimental Setup}
The experiment was carried out with Kaggle. The programming language used was Python. The configuration included a GPU P100 to expedite the training process of the deep learning model. This experiment used many Python libraries, including numpy (used for numerical and scientific computing), pandas (for data manipulation and analysis), os (for data read/write), tensorflow (for developing deep learning models), sklearn (used for XGBoost and performance evaluation), TensorFlow Lite (for model quantization), and matplotlib (for visualization).

\subsection{Result Analysis}
We performed a thorough assessment of the proposed solution for the classification of brain tumors. We employed multiple evaluation metrics for assessing the classifier, including accuracy, precision, recall, F1-score, and Matthews Correlation Coefficient (MCC). Among the four matrices, MCC is considered the most important one for measuring the model's completeness as it considers all the coordinates for the result calculation \cite{hicks2022evaluation}. These metrics serve as a foundation for our assessment and provide comprehensive insight into the capabilities of our model. We conducted a side-by-side comparison of MLP and XGBoost, two classification heads of the model. Moreover, we included confusion matrices and Receiver operating characteristic (ROC) curve to provide detailed insight into the performance of our model.

\begin{figure*}[h]
\centering
\begin{subfigure}{0.48\textwidth}
    \includegraphics[height=6cm, width=6.8cm]{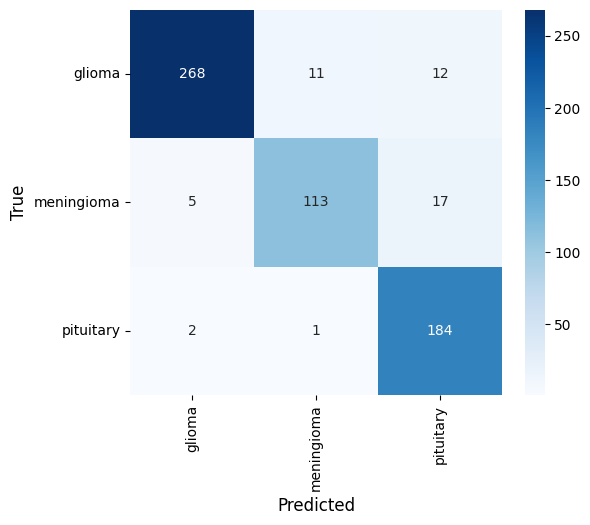}
    \caption{Confusion matrix of the fusion model with MLP head on Figshare dataset.}
    \label{fig:figshare-mlp-cm}
\end{subfigure}
\hfill
\begin{subfigure}{0.48\textwidth}
    \includegraphics[height=6cm, width=6.8cm]{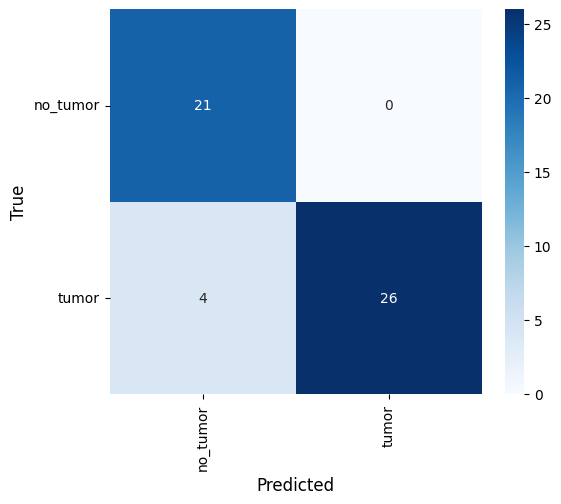}
    \caption{Confusion matrix of the fusion model with MLP head on Kaggle dataset.}
    \label{fig:kaggle-mlp-cm}
\end{subfigure}

\caption{Confusion matrix of the fusion model with MLP head on different datasets.}
\label{fig:mlp-cm}
\end{figure*}

Our initial classification head, the MLP, displayed remarkable outcomes in all metrics. It achieved a notable accuracy of 92.17\% and 92.16\% in the Figshare and the Kaggle dataset respectively, indicating that the predictions of our model were mostly accurate. The model's recall of 91.41\% on the Figshare and 92\% on the Kaggle dataset indicates its excellent performance in identifying true positive instances, while its precision of 91.40\% and 93.33\% on the Figshare and Kaggle dataset showcases its excellent performance in minimizing false positives. Impressively, the model achieved an F1-score of 91.20\% on the Figshare dataset and 92.08\% on the Kaggle dataset indicating a high level of balance between precision and recall. In addition, the proposed solution attained an MCC of 87.88\% and 85.32\% on the Figshare and the Kaggle datasets, respectively, suggesting a comprehensive solution that considers both true and false positives as well as true and false negatives.

\begin{figure*}[h]
\centering
\begin{subfigure}{.48\textwidth}
    \includegraphics[height=6cm, width=6.8cm]{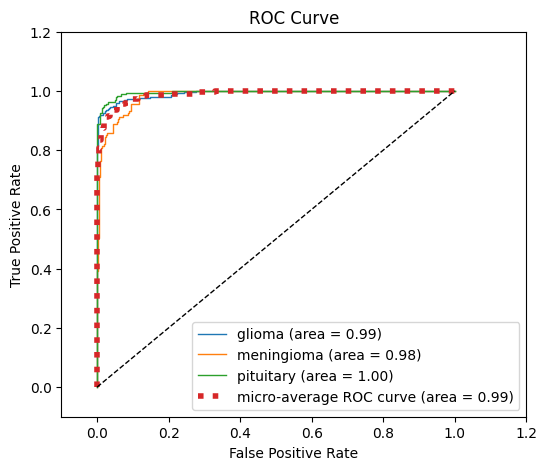}
    \caption{ROC curve of the fusion model with MLP head on Figshare dataset.}
    \label{fig:figshare-mlp-roc}
\end{subfigure}
\hfill
\begin{subfigure}{0.5\textwidth}
    \includegraphics[height=6cm, width=6.8cm]{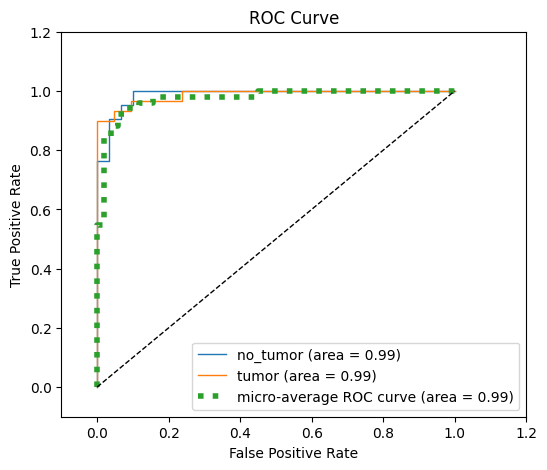}
    \caption{ROC curve of the fusion model with MLP head on Kaggle dataset.}
    \label{fig:kaggle-mlp-roc}
\end{subfigure}

\caption{ROC curve of the fusion model with MLP head on different datasets.}
\label{fig:mlp-roc}
\end{figure*}

However, with acceptable accomplishments, the MLP classification head highlighted some areas of concern. Notably, the confusion matrix on the Figshare dataset, shown in Figure \ref{fig:figshare-mlp-roc}, illustrates that the model struggles with inaccuracies that include 17 meningioma tumors incorrectly identified as pituitary tumors, 12 glioma cases incorrectly labeled as pituitary tumors, and 11 cases of glioma-type brain tumors misclassified as meningiomas. Likewise, Figure \ref{fig:kaggle-mlp-cm} indicates that the model misclassifies 4 tumorous images as non-tumorous.

\begin{figure*}[h]
\centering
\begin{subfigure}{0.48\textwidth}
    \includegraphics[height=6cm, width=6.8cm]{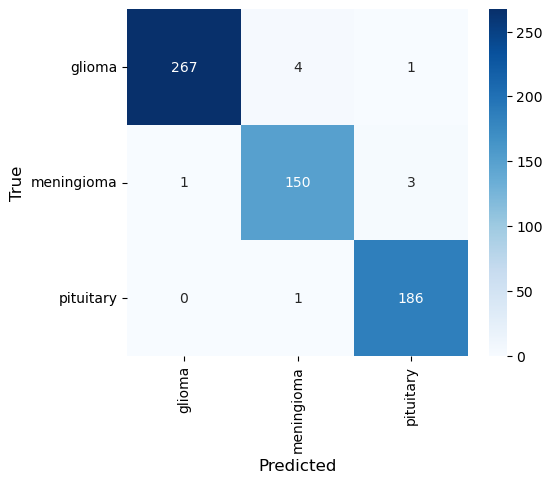}
    \caption{Confusion matrix of the fusion model with XGBoost head on Figshare dataset.}
    \label{fig:figshare-xgb-cm}
\end{subfigure}
\hfill
\begin{subfigure}{0.48\textwidth}
    \includegraphics[height=6cm, width=6.8cm]{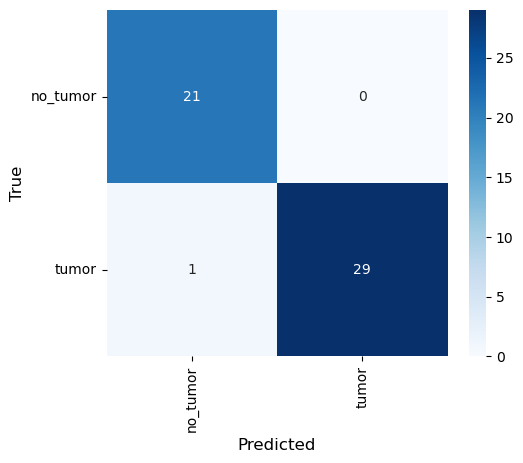}
    \caption{Confusion matrix of the fusion model with XGBoost head on Kaggle dataset.}
    \label{fig:kaggle-xgb-cm}
\end{subfigure}

\caption{Confusion matrix of the fusion model with XGBoost head on different datasets.}
\label{fig:xgb-cm}
\end{figure*}

Similarly, the ROC curve on the Figshare dataset, presented in Figure \ref{fig:figshare-mlp-roc}, shows that the pituitary class achieves a perfect value of 1.0. In contrast, glioma and meningioma achieved ROC values of 0.99 and 0.98 respectively, resulting in a micro-average ROC value of 0.99. Figure \ref{fig:kaggle-mlp-roc} presents the ROC curve of the model with MLP head on the Kaggle dataset. Both the tumor and non-tumor images achieve an ROC value of 0.99, resulting in a micro average accuracy of 0.99.

\begin{figure*}[h]
\centering
\begin{subfigure}{0.48\textwidth}
    \includegraphics[height=6cm, width=6.8cm]{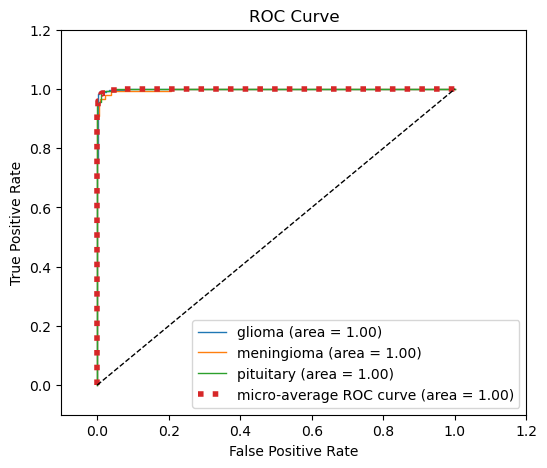}
    \caption{ROC curve of the fusion model with XGBoost head on Figshare dataset.}
    \label{fig:figshare-xgb-roc}
\end{subfigure}
\hfill
\begin{subfigure}{0.48\textwidth}
    \includegraphics[height=6cm, width=6.8cm]{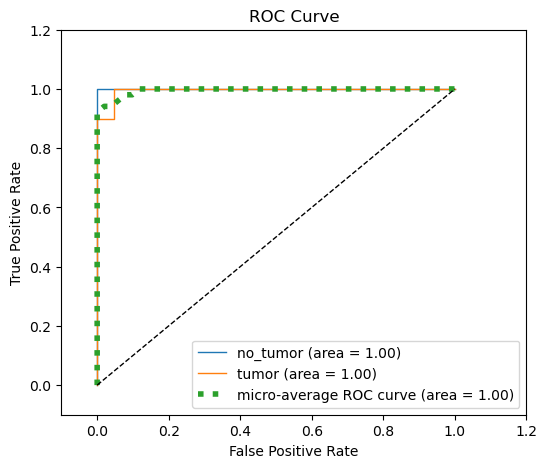}
    \caption{ROC curve of the fusion model with XGBoost head on Kaggle dataset.}
    \label{fig:kaggle-xgb-roc}
\end{subfigure}

\caption{ROC curve of the fusion model with XGBoost head on different datasets.}
\label{fig:xgb-roc}
\end{figure*}

The insertion of the XGBoost classification head significantly enhanced the performance of the model. On the Fighsare dataset, it demonstrated astounding performance as seen by its accuracy of 98.36\%, precision rate of 98.34\%, and recall value of 98.09\%. The F1-score has raised to 98.21\%. Additionally, a remarkable MCC score of 97.48\% indicates a high distinguishing ability of the model. The remarkable capacity of the XGBoost classification head to minimize large errors considerably makes it stand out. The confusion matrix presented in Figure \ref{fig:figshare-xgb-cm} shows only four glioma-type brain tumors misclassified as meningiomas and only three meningioma tumors misclassified as pituitary tumors. In addition, none of the remaining coefficients exhibited more than one misclassification. The robustness of the model in distinguishing between true positive and false positive instances was further demonstrated by the ROC curve, confirming its great discriminatory capacity. Figure \ref{fig:figshare-xgb-roc} shows that the model with the XGBoost classification head achieves a perfect ROC value of 1.0 in all classes. 

Likewise, the model achieves noteworthy results on the Kaggle dataset with accuracy, precision, and recall values of 98.04\%, 98.33\%, and 97.73\% respectively. Moreover, an F1-score of 97.99\% and and MCC value of 96.06\% further testifies it's classification ability. Figure \ref{fig:kaggle-xgb-cm} presents the confusion matrix of the model on the Kaggle dataset. The figure exhibits only one misclassification among 51 instances. The ROC curve, presented in Figure \ref{fig:kaggle-xgb-roc}, presents the model achieves a perfect ROC value of 1.0 in all classes.

\subsection{Model Attention}
This work employed Gradient-weighted Class Activation Mapping (Grad-CAM) \cite{selvaraju2017grad} to investigate the model's priority regions during decision-making. Grad-CAM is an effective approach that enhances the transparency of the decision-making process in deep learning models, specifically in the domain of computer vision. It offers a heatmap of the specific regions of input images that have a major contribution on a model's predictions. Grad-CAM utilizes the gradient values of the target class with respect to the feature maps in the last convolutional layer of a neural network. This technique generates a heatmap that visually represents the relative importance of various regions within an input image in relation to the model's decision. This research used Grad-CAM to render the preferences of our intricate neural network.

\begin{figure}[h]
    \centering
    \includegraphics[height=8cm,width=0.7\linewidth]{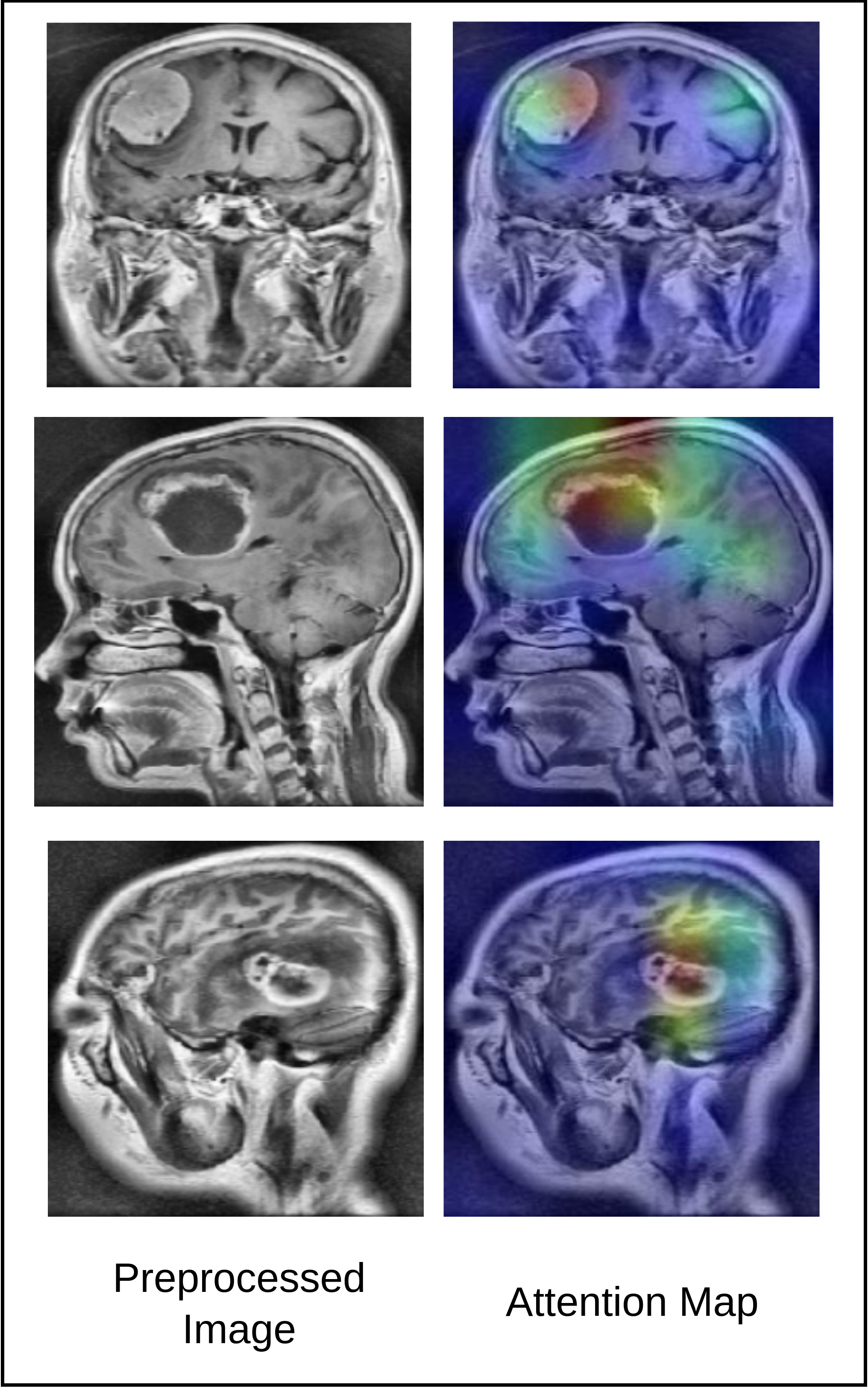}
    \caption{Attention map of the model on different brain tumor images. The bright colors on the attention map indicate focus areas.}
    \label{fig:xai}
\end{figure}

Figure \ref{fig:xai} showcases a side-by-side comparison of three preprocessed images along with their corresponding heatmaps on the Fighsare dataset. The attention map illustrates that the model predominantly focuses on the tumor regions during the decision-making process. While our approach primarily caters to critical regions, our investigation also reveals cases where it incorrectly concentrates on unimportant regions.

\subsection{Comparison with Existing Works}
Over the past few years, several studies have been conducted to classify brain tumors. Most studies have employed CNN as the classifier. The authors predominantly employed various pretrained feature extractors that were slightly fine-tuned for the final classification. Custom shallow CNN architectures without transfer learning have also achieved acceptable performance. Additionally, some researchers have localized the tumor portion leveraging R-CNN and employed various deep learning algorithms for final classification. Notably, a combination of deep learning and traditional machine learning is also seen which has produced an outstanding performance. A comparison with recent studies on the Figshare dataset is presented in Table \ref{tab:comp}. 

\begin{table*}[h]
    \centering
    \caption{Performance comparison with existing works on the Figshare dataset.}
    \begin{tabular}{p{2.2cm} p{3.6cm} p{1.6cm} p{1.5cm} p{1.5cm} p{1.5cm}}
    \hline
        \textbf{Paper} & \textbf{Method} & \textbf{Accuracy} & \textbf{Precision} & \textbf{Recall} & \textbf{F1-score} \\ \hline
        Khan et al. \cite{khan2022accurate} & 23 Layer custom CNN architecture & 0.978 & 0.965 & 0.964 & 0.964 \\ 
        Masood et al. \cite{masood2021novel} & DenseNet-41-based Mask-RCNN & 0.963 & - & - & - \\ 
        Diaz et al. \cite{diaz2021deep} & Three-stream CNN architecture & 0.973 & 0.9 & - & - \\ 
        Sultan et al. \cite{sultan2019multi} & Custom CNN with cross channel normalization & 0.961 & 0.963 & 0.944 & 0.953 \\ 
        Swati et al. \cite{swati2019brain} & Pretrained VGG19 & 0.9482 & 8734 & 9681 & 0.918 \\ 
        Ahmad et al. \cite{ahmad2022brain} & Resnet50 & 0.9625 & 0.833 & 0.769 & 0.8 \\ 
        Haq et al. \cite{haq2022hybrid} & Hybrid CNN and SVM method & 0.98 & 0.967 & 0.971 & 0.969 \\ 
        Proposed & Resnet152V2 and VGG16 fusion model with XGBoost & 0.9836 & 0.9834 & 0.98 & 0.982 \\ \hline
    \end{tabular}
    \label{tab:comp}
\end{table*}

Likewise, Table \ref{tab:comp-kaggle} holds a direct comparison with the recent research on the Kaggle dataset. Since the dataset has only 253 instances, lightweight neural networks, such as ResNet50, often produce an acceptable accuracy. The comparison shows that the proposed model outperforms most existing methods by a noticeable margin. 

\begin{table}[h]
    \centering
    \caption{Performance comparison with existing works on the Kaggle dataset.}
    \begin{tabular}{p{2.2cm} p{3.6cm} p{1.6cm} p{1.5cm} p{1.5cm} p{1.5cm}}
    \hline
        Paper & Method & Accuracy & Precision  & Recall & F1-Score \\ \hline
        Rahman et al. \cite{rahman2023mri} & Parralel custom CNN & 0.9733 & 0.98 & 0.9750 & 0.9750 \\ 
        Saxena et al.  \cite{saxena2020predictive} & ResNet50 & 0.9500 & ~ & ~ & 0.952 \\ 
        Ccinar et al. \cite{ccinar2020detection} & modified Resnet50  & 0.9707 & - & 89.74 & 93.33 \\ 
        Alsaif et al. \cite{alsaif2022novel} & VGG16 & 0.9600 & 0.93 & 1 & 0.97 \\ 
        Asif et al. \cite{asif2022improving} & Xception & 0.9194 & - & 0.9655 & 0.918 \\ 
        Moirangthem et al. \cite{moirangthem2021image} & ResNet 50 & 0.8718 & 0.8847 & 0.8525 & 0.8693 \\ 
        Proposed & Resnet152V2 and VGG16 fusion model with XGBoost & 0.9804 & 0.9833 & 0.9773 & 0.9606 \\ \hline
    \end{tabular}
    \label{tab:comp-kaggle}
\end{table}

The existing solutions have three major drawbacks that the research addresses. Firstly, the medical images typically consist of noises and uneven lighting which significantly influences the classification performance. Therefore, the proposed system includes a variety of data preparation techniques to improve the image quality, addressing a common shortcoming of recent efforts in the field. Moreover, some deep neural networks such as VGG16 drop performance due to the vanishing gradient problem. We also presented a solution that addresses this issue without compromising performance and resource consumption. Additionally, deep learning architectures tend to produce relatively low performance if the network is not pre-trained. In the transfer learning method, the top layer is typically trained from scratch which may lead to a decrease in the performance if the training data is limited. To address this shortcoming, we replaced the top MLP layer with XGBoost, a high-performing machine learning classifier. With all the improvements, the proposed solution achieved significant performance in brain tumor detection.

\subsection{Discussion}
The study on a computer vision based diagnosis of brain tumors addresses the critical need for accurate and fast diagnosis of the disease. While existing studies offer solution to the problem, they primarily focus on designing classifiers with high accuracy. However, since deep learning models typically have a large size, they are infeasible to integrate in remote areas, particularly where high-configuration devices are unavailable. We have addressed this limitation by carefully designing the classifier to be lightweight without compromising the performance. Furthermore, we have leveraged quantization to effectively reduce the model size to 73.881 MB which makes the model easily deployable into edge devices. Additionally, some research employs only one dataset for evaluating the classifier. This increases the threat of the model being biased towards a particular dataset. To address the censorious issue, we have employed Grad-CAM to visualize the attention heatmap of the model. According to the analysis, the model predominantly focuses on the tumor regions for making the decision. The experiment enhances the reliability of the classifier. The proposed system, however, lacks additional studies that can be addressed in the future. Firstly, due to the variety in the image acquisition process of different medical devices, the model needs to be domain-adaptive for its widespread utilization. Future studies may include additional datasets, acquired from different devices to evaluate the model's efficiency. Additionally, no ablation study is performed to understand the impact of the hyperparameters. A thorough ablation study will provide further insights into the model which may result in an advancement in the classification accuracy. Nevertheless, considering the high classification accuracy and resource efficiency, the solution can be integrated into diagnostic centers located in remote regions with poor internet connectivity and unavailability of high computing devices.

% The presented evaluation matrics along with a thorough comparison with the existing solutions indicate the model's efficiency in classifying brain tumors. The noteworthy performance can be attributed to two major factors. Firstly, the proposed four-stage image preprocessing technique significantly improves the image quality. While the ROI selection process effectively eliminates unnecessary backgrounds, CLAHE enhances the contrast and visibility of the tumor images. Furthermore, the data augmentation technique creates new images, leveraging six augmentation algorithms to prevent overfitting. Finally, resizing ensures the images are compatible with the pretrained classifiers. Secondly, the carefully designed architecture of the fusion model ensures the model receives features from multiple streams, resulting in more accurate decision-making. Due to the vanishing gradient problem, commonly observed in deep neural networks, some fine features get lost during the feature extraction process. These features play a vital role in the decision-making process. The proposed architecture is designed to address the limitations, which resulted in a high classification accuracy. Moreover, we have leveraged Grad-CAM to determine the regions where the model is focusing during inference.

\section{Conclusion}
\label{conclusion}
This study introduces a novel method designed to classify brain tumors in an efficient manner. The classification process encompasses a range of image preprocessing methods aimed at enhancing image quality. For classification, we have proposed a deep learning fusion model comprising ResNet152V2 and VGG16. Due to the over-smoothing problem of VGG16, we have altered the architecture to preserve the fine gradients. Following a small fine-tuning, the retrieved features are concatenated and finally classified using XGBoost. The suggested lightweight system is designed to enhance both performance and resource efficiency, attaining an accuracy of over 98\% while using merely 2.8 million trainable parameters and undergoing 50 training epochs. The model is further compressed through 8-bit quantization, making the model easily deployable in edge devices. The model is thoroughly evaluated on the Figshare and the Kaggle datasets, and the findings demonstrated its superiority over existing solutions. The solution, however, has two major drawbacks. Firstly, the solution is evaluated on only two datasets. Medical images may exhibit slight variations due to the differences in the capturing devices. Domain adaptation can serve as a potential solution to this problem. Moreover, this research lacks an ablation study that may have presented more insights into this research. Further studies can be conducted to address these limitations.

% \subsection*{Author Contributions} 
% Describe contributions of each author to the paper, using the first initial and full last name. 

% \medskip Examples:

% ``S. Zhang conceived the idea and designed the experiments.''

% ``E. F. Mustermann and J. F. Smith conducted the experiments.''

\section*{Data Availability}
The Figshare dataset is available at \url{https://figshare.com/articles/dataset/brain_tumor_dataset/1512427} (accessed on 16 May 2024) and the Kaggle dataset is available at: \url{https://www.kaggle.com/datasets/navoneel/brain-mri-images-for-brain-tumor-detection/} (accessed on 16 May 2024).

\section*{Conflicts of Interest}
The authors declare no conflict of interest.

\printbibliography

@article{sultan2019multi,
  title={Multi-classification of brain tumor images using deep neural network},
  author={Sultan, Hossam H and Salem, Nancy M and Al-Atabany, Walid},
  journal={IEEE access},
  volume={7},
  pages={69215--69225},
  year={2019},
  publisher={IEEE}
}

@article{kokkalla2021three,
  title={Three-class brain tumor classification using deep dense inception residual network},
  author={Kokkalla, Srinath and Kakarla, Jagadeesh and Venkateswarlu, Isunuri B and Singh, Munesh},
  journal={Soft Computing},
  volume={25},
  pages={8721--8729},
  year={2021},
  publisher={Springer}
}

@article{islam2023toward,
  title={Toward Lightweight Diabetic Retinopathy Classification: A Knowledge Distillation Approach for Resource-Constrained Settings},
  author={Islam, Niful and Jony, Md Mehedi Hasan and Hasan, Emam and Sutradhar, Sunny and Rahman, Atikur and Islam, Md Motaharul},
  journal={Applied Sciences},
  volume={13},
  number={22},
  pages={12397},
  year={2023},
  publisher={MDPI}
}

@article{magdy2023performance,
  title={Performance Enhancement of Skin Cancer Classification using Computer Vision},
  author={Magdy, Ahmed and Hussein, Hadeer and Abdel-Kader, Rehab F and Abd El Salam, Khaled},
  journal={IEEE Access},
  year={2023},
  publisher={IEEE}
}

@inproceedings{deng2009imagenet,
  title={Imagenet: A large-scale hierarchical image database},
  author={Deng, Jia and Dong, Wei and Socher, Richard and Li, Li-Jia and Li, Kai and Fei-Fei, Li},
  booktitle={2009 IEEE conference on computer vision and pattern recognition},
  pages={248--255},
  year={2009},
  organization={Ieee}
}

@article{irmak2021multi,
  title={Multi-classification of brain tumor MRI images using deep convolutional neural network with fully optimized framework},
  author={Irmak, Emrah},
  journal={Iranian Journal of Science and Technology, Transactions of Electrical Engineering},
  volume={45},
  number={3},
  pages={1015--1036},
  year={2021},
  publisher={Springer}
}

@article{khan2022accurate,
  title={Accurate brain tumor detection using deep convolutional neural network},
  author={Khan, Md Saikat Islam and Rahman, Anichur and Debnath, Tanoy and Karim, Md Razaul and Nasir, Mostofa Kamal and Band, Shahab S and Mosavi, Amir and Dehzangi, Iman},
  journal={Computational and Structural Biotechnology Journal},
  volume={20},
  pages={4733--4745},
  year={2022},
  publisher={Elsevier}
}

@article{asif2023enhanced,
  title={An enhanced deep learning method for multi-class brain tumor classification using deep transfer learning},
  author={Asif, Sohaib and Zhao, Ming and Tang, Fengxiao and Zhu, Yusen},
  journal={Multimedia Tools and Applications},
  pages={1--28},
  year={2023},
  publisher={Springer}
}

@article{kumar2023human,
  title={Human brain tumor classification and segmentation using CNN},
  author={Kumar, Sunil and Kumar, Dilip},
  journal={Multimedia Tools and Applications},
  volume={82},
  number={5},
  pages={7599--7620},
  year={2023},
  publisher={Springer}
}

@article{kurdi2023brain,
  title={Brain Tumor Classification Using Meta-Heuristic Optimized Convolutional Neural Networks},
  author={Kurdi, Sarah Zuhair and Ali, Mohammed Hasan and Jaber, Mustafa Musa and Saba, Tanzila and Rehman, Amjad and Dama{\v{s}}evi{\v{c}}ius, Robertas},
  journal={Journal of Personalized Medicine},
  volume={13},
  number={2},
  pages={181},
  year={2023},
  publisher={MDPI}
}

@article{athisayamani2023feature,
  title={Feature Extraction Using a Residual Deep Convolutional Neural Network (ResNet-152) and Optimized Feature Dimension Reduction for MRI Brain Tumor Classification},
  author={Athisayamani, Suganya and Antonyswamy, Robert Singh and Sarveshwaran, Velliangiri and Almeshari, Meshari and Alzamil, Yasser and Ravi, Vinayakumar},
  journal={Diagnostics},
  volume={13},
  number={4},
  pages={668},
  year={2023},
  publisher={MDPI}
}

@article{vankdothu2022brain,
  title={Brain tumor MRI images identification and classification based on the recurrent convolutional neural network},
  author={Vankdothu, Ramdas and Hameed, Mohd Abdul},
  journal={Measurement: Sensors},
  volume={24},
  pages={100412},
  year={2022},
  publisher={Elsevier}
}

@article{noreen2020deep,
  title={A deep learning model based on concatenation approach for the diagnosis of brain tumor},
  author={Noreen, Neelum and Palaniappan, Sellappan and Qayyum, Abdul and Ahmad, Iftikhar and Imran, Muhammad and Shoaib, Muhammad},
  journal={IEEE Access},
  volume={8},
  pages={55135--55144},
  year={2020},
  publisher={IEEE}
}

@article{mohsen2023brain,
  title={Brain Tumor Classification Using Hybrid Single Image Super-Resolution Technique with ResNext101\_32x8d and VGG19 Pre-Trained Models},
  author={Mohsen, Saeed and Ali, Anas M and El-Rabaie, El-Sayed M and Elkaseer, Ahmed and Scholz, Steffen G and Hassan, Ashraf Mohamed Ali},
  journal={IEEE Access},
  year={2023},
  publisher={IEEE}
}

@article{shah2022robust,
  title={A robust approach for brain tumor detection in magnetic resonance images using finetuned efficientnet},
  author={Shah, Hasnain Ali and Saeed, Faisal and Yun, Sangseok and Park, Jun-Hyun and Paul, Anand and Kang, Jae-Mo},
  journal={IEEE Access},
  volume={10},
  pages={65426--65438},
  year={2022},
  publisher={IEEE}
}

@article{alsaif2022novel,
  title={A novel data augmentation-based brain tumor detection using convolutional neural network},
  author={Alsaif, Haitham and Guesmi, Ramzi and Alshammari, Badr M and Hamrouni, Tarek and Guesmi, Tawfik and Alzamil, Ahmed and Belguesmi, Lamia},
  journal={Applied sciences},
  volume={12},
  number={8},
  pages={3773},
  year={2022},
  publisher={MDPI}
}

@article{montaha2022timedistributed,
  title={Timedistributed-cnn-lstm: A hybrid approach combining cnn and lstm to classify brain tumor on 3d mri scans performing ablation study},
  author={Montaha, Sidratul and Azam, Sami and Rafid, AKM Rakibul Haque and Hasan, Md Zahid and Karim, Asif and Islam, Ashraful},
  journal={IEEE Access},
  volume={10},
  pages={60039--60059},
  year={2022},
  publisher={IEEE}
}

@article{musallam2022new,
  title={A new convolutional neural network architecture for automatic detection of brain tumors in magnetic resonance imaging images},
  author={Musallam, Ahmed S and Sherif, Ahmed S and Hussein, Mohamed K},
  journal={IEEE access},
  volume={10},
  pages={2775--2782},
  year={2022},
  publisher={IEEE}
}

@article{amin2020brain,
  title={Brain tumor classification based on DWT fusion of MRI sequences using convolutional neural network},
  author={Amin, Javaria and Sharif, Muhammad and Gul, Nadia and Yasmin, Mussarat and Shad, Shafqat Ali},
  journal={Pattern Recognition Letters},
  volume={129},
  pages={115--122},
  year={2020},
  publisher={Elsevier}
}

@article{khairandish2022hybrid,
  title={A hybrid CNN-SVM threshold segmentation approach for tumor detection and classification of MRI brain images},
  author={Khairandish, Mohammad Omid and Sharma, Meenakshi and Jain, Vishal and Chatterjee, Jyotir Moy and Jhanjhi, NZ},
  journal={Irbm},
  volume={43},
  number={4},
  pages={290--299},
  year={2022},
  publisher={Elsevier}
}

@article{kang2021mri,
  title={Mri-based brain tumor classification using ensemble of deep features and machine learning classifiers},
  author={Kang, Jaeyong and Ullah, Zahid and Gwak, Jeonghwan},
  journal={Sensors},
  volume={21},
  number={6},
  pages={2222},
  year={2021},
  publisher={MDPI}
}

@inproceedings{diaz2021deep,
  title={A deep learning approach for brain tumor classification and segmentation using a multiscale convolutional neural network},
  author={D{\'\i}az-Pernas, Francisco Javier and Mart{\'\i}nez-Zarzuela, Mario and Ant{\'o}n-Rodr{\'\i}guez, M{\'\i}riam and Gonz{\'a}lez-Ortega, David},
  booktitle={Healthcare},
  volume={9},
  pages={153},
  year={2021},
  organization={MDPI}
}

@inproceedings{liu2021swin,
  title={Swin transformer: Hierarchical vision transformer using shifted windows},
  author={Liu, Ze and Lin, Yutong and Cao, Yue and Hu, Han and Wei, Yixuan and Zhang, Zheng and Lin, Stephen and Guo, Baining},
  booktitle={Proceedings of the IEEE/CVF international conference on computer vision},
  pages={10012--10022},
  year={2021}
}

@inproceedings{wang2021pyramid,
  title={Pyramid vision transformer: A versatile backbone for dense prediction without convolutions},
  author={Wang, Wenhai and Xie, Enze and Li, Xiang and Fan, Deng-Ping and Song, Kaitao and Liang, Ding and Lu, Tong and Luo, Ping and Shao, Ling},
  booktitle={Proceedings of the IEEE/CVF international conference on computer vision},
  pages={568--578},
  year={2021}
}

@article{tummala2022classification,
  title={Classification of brain tumor from magnetic resonance imaging using vision transformers ensembling},
  author={Tummala, Sudhakar and Kadry, Seifedine and Bukhari, Syed Ahmad Chan and Rauf, Hafiz Tayyab},
  journal={Current Oncology},
  volume={29},
  number={10},
  pages={7498--7511},
  year={2022},
  publisher={MDPI}
}

@article{ferdous2023lcdeit,
  title={LCDEiT: A Linear Complexity Data-Efficient Image Transformer for MRI Brain Tumor Classification},
  author={Ferdous, Gazi Jannatul and Sathi, Khaleda Akhter and Hossain, Md Azad and Hoque, Mohammed Moshiul and Dewan, M Ali Akber},
  journal={IEEE Access},
  volume={11},
  pages={20337--20350},
  year={2023},
  publisher={IEEE}
}

@article{aloraini2023combining,
  title={Combining the Transformer and Convolution for Effective Brain Tumor Classification Using MRI Images},
  author={Aloraini, Mohammed and Khan, Asma and Aladhadh, Suliman and Habib, Shabana and Alsharekh, Mohammed F and Islam, Muhammad},
  journal={Applied Sciences},
  volume={13},
  number={6},
  pages={3680},
  year={2023},
  publisher={MDPI}
}

@article{Cheng2017,
  title={Enhanced performance of brain tumor classification via tumor region augmentation and partition},
  author={Cheng, Jun and Huang, Wei and Cao, Shuangliang and Yang, Ru and Yang, Wei and Yun, Zhaoqiang and Wang, Zhijian and Feng, Qianjin},
  journal={PloS one},
  volume={10},
  number={10},
  pages={e0140381},
  year={2015},
  publisher={Public Library of Science San Francisco, CA USA}
}

@article{nyo2022otsu,
  title={Otsu’s thresholding technique for MRI image brain tumor segmentation},
  author={Nyo, Myat Thet and Mebarek-Oudina, F and Hlaing, Su Su and Khan, Nadeem A},
  journal={Multimedia tools and applications},
  volume={81},
  number={30},
  pages={43837--43849},
  year={2022},
  publisher={Springer}
}

@inproceedings{he2016deep,
  title={Deep residual learning for image recognition},
  author={He, Kaiming and Zhang, Xiangyu and Ren, Shaoqing and Sun, Jian},
  booktitle={Proceedings of the IEEE conference on computer vision and pattern recognition},
  pages={770--778},
  year={2016}
}

@article{hwooi2022deep,
  title={Deep learning-based approach for continuous affect prediction from facial expression images in valence-arousal space},
  author={Hwooi, Stephen Khor Wen and Othmani, Alice and Sabri, Aznul Qalid Md},
  journal={IEEE Access},
  volume={10},
  pages={96053--96065},
  year={2022},
  publisher={IEEE}
}

@inproceedings{wang2018non,
  title={Non-local neural networks},
  author={Wang, Xiaolong and Girshick, Ross and Gupta, Abhinav and He, Kaiming},
  booktitle={Proceedings of the IEEE conference on computer vision and pattern recognition},
  pages={7794--7803},
  year={2018}
}

@article{kong2022classification,
  title={Classification and detection of COVID-19 X-Ray images based on DenseNet and VGG16 feature fusion},
  author={Kong, Lingzhi and Cheng, Jinyong},
  journal={Biomedical Signal Processing and Control},
  volume={77},
  pages={103772},
  year={2022},
  publisher={Elsevier}
}

@inproceedings{fu2019dual,
  title={Dual attention network for scene segmentation},
  author={Fu, Jun and Liu, Jing and Tian, Haijie and Li, Yong and Bao, Yongjun and Fang, Zhiwei and Lu, Hanqing},
  booktitle={Proceedings of the IEEE/CVF conference on computer vision and pattern recognition},
  pages={3146--3154},
  year={2019}
}

@article{hicks2022evaluation,
  title={On evaluation metrics for medical applications of artificial intelligence},
  author={Hicks, Steven A and Str{\"u}mke, Inga and Thambawita, Vajira and Hammou, Malek and Riegler, Michael A and Halvorsen, P{\aa}l and Parasa, Sravanthi},
  journal={Scientific reports},
  volume={12},
  number={1},
  pages={5979},
  year={2022},
  publisher={Nature Publishing Group UK London}
}

@inproceedings{selvaraju2017grad,
  title={Grad-cam: Visual explanations from deep networks via gradient-based localization},
  author={Selvaraju, Ramprasaath R and Cogswell, Michael and Das, Abhishek and Vedantam, Ramakrishna and Parikh, Devi and Batra, Dhruv},
  booktitle={Proceedings of the IEEE international conference on computer vision},
  pages={618--626},
  year={2017}
}

@misc{kaggle_brain_mri,
  author       = {Navoneel Chakrabarty},
  title        = {Brain MRI Images for Brain Tumor Detection},
  year         = {2019},
  howpublished = {\url{https://www.kaggle.com/datasets/navoneel/brain-mri-images-for-brain-tumor-detection/}},
  note         = {(Accessed on May 16, 2024)}
}

@article{rahman2023mri,
  title={MRI brain tumor detection and classification using parallel deep convolutional neural networks},
  author={Rahman, Takowa and Islam, Md Saiful},
  journal={Measurement: Sensors},
  volume={26},
  pages={100694},
  year={2023},
  publisher={Elsevier}
}

@incollection{saxena2020predictive,
  title={Predictive modeling of brain tumor: a deep learning approach},
  author={Saxena, Priyansh and Maheshwari, Akshat and Maheshwari, Saumil},
  booktitle={Innovations in Computational Intelligence and Computer Vision: Proceedings of ICICV 2020},
  pages={275--285},
  year={2020},
  publisher={Springer}
}

@article{ccinar2020detection,
  title={Detection of tumors on brain MRI images using the hybrid convolutional neural network architecture},
  author={{\c{C}}inar, Ahmet and Yildirim, Muhammed},
  journal={Medical hypotheses},
  volume={139},
  pages={109684},
  year={2020},
  publisher={Elsevier}
}

@ARTICLE{asif2022improving,
  author={Asif, Sohaib and Yi, Wenhui and Ain, Qurrat Ul and Hou, Jin and Yi, Tao and Si, Jinhai},
  journal={IEEE Access}, 
  title={Improving Effectiveness of Different Deep Transfer Learning-Based Models for Detecting Brain Tumors From MR Images}, 
  year={2022},
  volume={10},
  number={},
  pages={34716-34730},
  doi={10.1109/ACCESS.2022.3153306}
}

@inproceedings{moirangthem2021image,
  title={Image classification and retrieval framework for Brain Tumour Detection using CNN on ROI segmented MRI images},
  author={Moirangthem, Manganleima and Singh, Thounaojam Rupachandra and Singh, Th Tangkeshwar},
  booktitle={2021 5th International Conference on Electrical, Electronics, Communication, Computer Technologies and Optimization Techniques (ICEECCOT)},
  pages={696--700},
  year={2021},
  organization={IEEE}
}

@article{masood2021novel,
  title={A novel deep learning method for recognition and classification of brain tumors from MRI images},
  author={Masood, Momina and Nazir, Tahira and Nawaz, Marriam and Mehmood, Awais and Rashid, Junaid and Kwon, Hyuk-Yoon and Mahmood, Toqeer and Hussain, Amir},
  journal={Diagnostics},
  volume={11},
  number={5},
  pages={744},
  year={2021},
  publisher={MDPI}
}

@article{swati2019brain,
  title={Brain tumor classification for MR images using transfer learning and fine-tuning},
  author={Swati, Zar Nawab Khan and Zhao, Qinghua and Kabir, Muhammad and Ali, Farman and Ali, Zakir and Ahmed, Saeed and Lu, Jianfeng},
  journal={Computerized Medical Imaging and Graphics},
  volume={75},
  pages={34--46},
  year={2019},
  publisher={Elsevier}
}

@article{ahmad2022brain,
  title={Brain tumor classification using a combination of variational autoencoders and generative adversarial networks},
  author={Ahmad, Bilal and Sun, Jun and You, Qi and Palade, Vasile and Mao, Zhongjie},
  journal={Biomedicines},
  volume={10},
  number={2},
  pages={223},
  year={2022},
  publisher={MDPI}
}

@article{haq2022hybrid,
  title={A hybrid approach based on deep cnn and machine learning classifiers for the tumor segmentation and classification in brain MRI},
  author={Haq, Ejaz Ul and Jianjun, Huang and Huarong, Xu and Li, Kang and Weng, Lifen},
  journal={Computational and Mathematical Methods in Medicine},
  volume={2022},
  year={2022},
  publisher={Hindawi}
}
\end{document}